\documentclass[journal,twoside,web]{ieeecolor}
\usepackage{tmi}
\usepackage{cite}
\usepackage{amsmath,amssymb,amsfonts}
\usepackage{algorithmic}
\usepackage{graphicx}
\usepackage{subfigure}
\usepackage{textcomp}
\usepackage{multirow}
\usepackage{booktabs}
\usepackage{bm}
\usepackage{bbding}
\usepackage{amssymb}

\usepackage{hyperref}
\hypersetup{hidelinks}
\usepackage{amsmath}

\usepackage{algorithm}
\usepackage{algorithmic}

\makeatletter
\newcommand{\rmnum}[1]{\romannumeral #1}
\newcommand{\Rmnum}[1]{\expandafter\@slowromancap\romannumeral #1@}
\makeatother

\usepackage[T1]{fontenc} 
\begin{document}
	\title{A Dual-Attention Learning Network with Word and Sentence Embedding for Medical Visual Question Answering}
	\author{Xiaofei Huang, Hongfang Gong
	\thanks{This work was supported in part by the National Natural Science Foundation of China under Grant 61972055, and in part by the Natural Science Foundation of Hunan Province under Grant 2021JJ30734. (Corresponding author: Hongfang Gong.)}
	\thanks{Xiaofei Huang, Hongfnag Gong are with the school of Mathematics and Statistics, Changsha University of Science and Technology, Changsha 410114, China (e-mail:xiaofeihuang2021@163.com; ghongfang@126.com).}
}
\maketitle

\begin{abstract}
Research in medical visual question answering (MVQA) can contribute to the development of computer-aided diagnosis. MVQA is a task that aims to predict accurate and convincing answers based on given medical images and associated natural language questions. This task requires extracting medical knowledge-rich feature content and making fine-grained understandings of them. Therefore, constructing an effective feature extraction and understanding scheme are keys to modeling. Existing MVQA question extraction schemes mainly focus on word information, ignoring medical information in the text. Meanwhile, some visual and textual feature understanding schemes cannot effectively capture the correlation between regions and keywords for reasonable visual reasoning. In this study, a dual-attention learning network with word and sentence embedding (WSDAN) is proposed. We design a module, transformer with sentence embedding (TSE), to extract a double embedding representation of questions containing keywords and medical information. A dual-attention learning (DAL) module consisting of self-attention and guided attention is proposed to model intensive intramodal and intermodal interactions. With multiple DAL modules (DALs), learning visual and textual co-attention can increase the granularity of understanding and improve visual reasoning. Experimental results on the ImageCLEF 2019 VQA-MED (VQA-MED 2019) and VQA-RAD datasets demonstrate that our proposed method outperforms previous state-of-the-art methods. According to the ablation studies and Grad-CAM maps, WSDAN can extract rich textual information and has strong visual reasoning ability.

\end{abstract}

\begin{IEEEkeywords}
Medical visual question answering, double embedding, medical information, guided attention, visual reasoning.
\end{IEEEkeywords}

\section{Introduction}
\label{sec:introduction}
Owing to intensive research in computer vision (CV) and natural language processing (NLP), related multimodal learning tasks, such as automatic image annotation\cite{zd}, video question answering\cite{spqa}, cross-modal information retrieval\cite{mhtn}, and visual question answering (VQA)\cite{2015vqa}\cite{VQAreview}, have attracted great interest. Medical visual question answering (MVQA) is a socially significant application of VQA and one of the current research hotspots in computer-aided diagnosis (CAD) technology. A mature MVQA system can relieve the burden on medical staff, provide them with valuable second opinions on medical images, and reduce the risk of misdiagnosis\cite{MVQA}.

MVQA requires us to answer relevant natural language (NL) questions in combination with given medical images. The complete modeling process can be described as follows. First, features are extracted from a given image and question. Second, the content of these features are understood and fused. Finally, the fused vectors are used to formulate the possible answers. Some research work has shown that because of the limitations of specialist medical concepts, challenges remain in understanding clinical texts\cite{CR}\cite{muvam}. Thus, text features that are rich in medical information must be extracted, but most studies have ignored this need. At the same time, previous research on feature fusion is imperfect, which hinders the model’s ability to perform correct visual reasoning and select core regions and keywords related to the answer. Yu \textit{et al.} proved through experiments that some dense interaction fusion schemes have low scalability\cite{MCAN}, and that schemes based on multi-head attention methods are rarely applied to MVQA\cite{MVQA}.

Word embedding, a technique for representing words as real-valued vectors, is commonly used in text processing. Most of the text extraction work in VQA and MVQA apply this technique to convert words into question features, thereby obtaining keyword information. The bi-branch model proposed by Liu \textit{et al.} uses the sum of token, position, and segment embedding as the question feature representation\cite{bpi}. However, many words have completely different public and medical meanings (e.g.,  "\textit{patient}" and "\textit{mass}"), so relying on words alone is not enough to gather sufficient medical information. Question-Centric Multimodal Low-rank Bilinear (QCMLB)\cite{QC-mlb} uses skip-thought vectors\cite{skipthoughts} to directly extract the sentence semantics of the question, which can obtain medical expertise but lose keyword information. As a result, the correct keywords and core areas may not be highlighted in the subsequent feature understanding process.

Multi-Modal Relation Attention Network (Mranet) obtains keyword information and semantic relations between words through word attention and self-guided relational attention, respectively. Then, the sum of the outputs of these two attention mechanisms is used as a question feature to answer complex queries\cite{mranet}. Yang \textit{et al.} added the question-type information after fusing multimodal features to narrow the candidate answer space\cite{cqta}. The aforementioned methods can obtain more valuable information from the text. However, because of the single question type and simple relationship between words in clinical NL questions, using these schemes does not sufficiently help in extracting medical information from text. Zhan \textit{et al.} designed Question-Conditioned Reasoning (QCR) and Type-Conditioned Reasoning (TCR) modules to extract close-ended and open-ended question features, respectively. Gupta \textit{et al.} generated word embedding and sub-word embedding, which were used with the integer sequence of the question to obtain text features\cite{hierarchical}. Although these schemes enrich text representation, they are more focused on the use of word information and still fall short in the semantic representation of medical questions. 

Attention mechanism\cite{attention} is not only successfully applied to unimodal tasks, but also plays an important role in many multimodal tasks like text matching\cite{bi} and visual captioning\cite{ca}. Recent studies have shown that learning co-attention for visual and textual elements at the same time can lead to a fine-grained understanding of the image and question, thereby improving the visual reasoning ability of the model and enabling more accurate predictions\cite{mfh}. Most works on co-attention focusing on MVQA are based on VQA, and each of these studies have flaws. Bilinear attention networks (BAN) achieved the purpose of simultaneously learning two modal attention distributions by adding an attention matrix to the bilinear model\cite{BAN}. Dense co-attention network (DCN) integrated visual and textual features by repeatedly interacting attention weights\cite{DCN}. Although BAN and DCN can achieve dense interaction of modalities, adding depth provides minimal improvement in performance. Of course, such schemes are beneficial to MVQA at present because the dataset is quite small, but as the amount of data increases, the effect and scalability of the model is affected. Modular Co-Attention Network (MCAN) proposed by Yu \textit{et al.} improves the model’s understanding of image content through the synergy of question-guided attention and self-attention\cite{MCAN}. However, it does not consider the image-guided attention and has not been applied to MVQA yet. BERT-based MVQA models, CGMVQA\cite{cgmvqa} and Multimodal Medical BERT (MMBERT)\cite{mmbert}, only use multi-head attention for intramodal and intermodal interactions. They are simple, but because they are learning two interactions at the same time, which weakens the effect of each, resulting in feature understanding and visual reasoning are inadequate.

In this study, we propose a Dual-Attention Learning Network with Word and Sentence Embedding (WSDAN) for MVQA to address the issues of inadequate extraction of medical information and poor granularity of feature understanding. Specifically, we design a module called transformer with sentence embedding (TSE) to extract question features. We used word and sentence embeddings to obtain keywords and medical information for the questions, respectively. A double embedding representation of the question is obtained through TSE. Then, we propose a dual-attention learning (DAL) module to learn self-attention and guided attention. The fusion component consisting of several DAL modules enhances the understanding of features and improves visual reasoning by learning co-attention between vision and text.

The present study’s contributions are summarized as follows: 
\begin{itemize}
\item A WSDAN is proposed for MVQA, which mainly solves the problems where the extracted question features lack medical information and does not completely understand the features; 
\item The TSE and DAL modules are designed. TSE extracts question features containing keywords and medical information. DAL models intramodal and intermodal interactions by learning self-attention and guided attention;
\item Our WSDAN achieves better performance on the \textit{VQA-MED 2019} and \textit{RAD} datasets. In addition, ablation experiments are performed to verify the effectiveness of the TSE module and determine whether DALs can improve feature understanding and enhance visual reasoning.
\end{itemize}

The rest of this paper is structured as follows. Section \Rmnum{2} briefly reviews related studies. Section \Rmnum{3} presents the entire modeling process and the various details of WSDAN. Section \Rmnum{4} describes the experiments and presents an analysis of the results. In section \Rmnum{5} the discussion and conlusion are provided. 

\section{Related Work}
Since ImageCLEF first hosted the visual question answering challenge in the medical domain\cite{clmvqa}, a growing number of researchers have participated in it and explored many methods for MVQA. Overall, the strategies used for MVQA and VQA tasks are similar, and both methods include four core learnable modules: (\rmnum{1}) a question encoder for extracting textual features, (\rmnum{2}) an image encoder for extracting visual features, (\rmnum{3}) a fusion algorithm for combining the extracted modal features, and (\rmnum{4}) a classifier that selects the optimal answer among a group of candidates.

\subsection{Question Encoder}
The purpose of this device is to encode the input questions, and Recurrent Neural Network (RNN) is ofter preferred for VQA question encoder because of the excellent performance of RNN in NLP. However, it cannot obtain the complete context because the auto-regressive structure is limited. The pre-trained Bidirectional Encoder Representations from Transformers (Bert) model proposed by Google adjusts the word vectors according to the context and obtains a context-rich word representation\cite{bert}. Although differences exist between the corpora in the general and medical domains, many MVQA studies have directly adopted pre-trained BERT to extract words as question features. 

In recent years, methods to create sentence embedding have led to a breakthrough in textual data representation. Ryan \textit{et al.} proposed skip-thought vectors to predict the context of the target sentence through an encoder–decoder composed of RNNs\cite{skipthoughts}. It can obtain vector representations and learn continuity relations between sentences but is less computationally efficient. Quick-thought vectors proposed by Lajanugen \textit{et al.} convert the predictive behavior of skip-thought vectors into categorical behavior, thereby improving sentence representation and computational efficiency\cite{quickthoughts}. Based on BERT, sentence BERT combines the capabilities of the Siamese and triplet network to create high-quality sentence representations\cite{sbert}. These methods are rarely used in MVQA.

\subsection{Image Encoder}
The image encoder is suitable for extracting visual features from image inputs, with pre-trained CNNs being the most common in VQA. The widely accepted CNNs are trained and tested on the ImageNet dataset, and researchers select suitable image encoders based on their performance. Visual Geometry Group Networks (VGGNet)\cite{vgg} and Residual Networks (ResNet)\cite{resnet} have been successfully applied to VQA and MVQA tasks. The main reason is that the features they extract are more general and effective for datasets other than ImageNet. Mixture of Enhanced Visual Features (MEVF) proposed by Nguyen \textit{et al.} used model-agnostic meta-learning (MAML) and Convolutional de-noising auto-encoder (CADE) to overcome the limitations of insufficient medical data and enhance visual features, which are extracted as image coding\cite{mevf}.

As related research continues to evolve, the images provided in the VQA task become more complex and natural. They require us to detect all objects of the input image for semantic segmentation. However, the rich features extracted by VGGNet and ResNet are specific to the image as a whole, which is more suitable for image classification. Gradually, CNN detection models (e.g., Faster Region-based Convolutional Network method (Faster R-CNN)\cite{ren2017faster} and Up-Down\cite{up-down}) are becoming the image encoder of choice in VQA. However, owing to the lack of a large-scale detection dataset in MVQA at present, most of the existing studies on image encoders are similar to the CNN classification model described above. 

\begin{figure*}[h]
\centerline{\includegraphics[width=16cm]{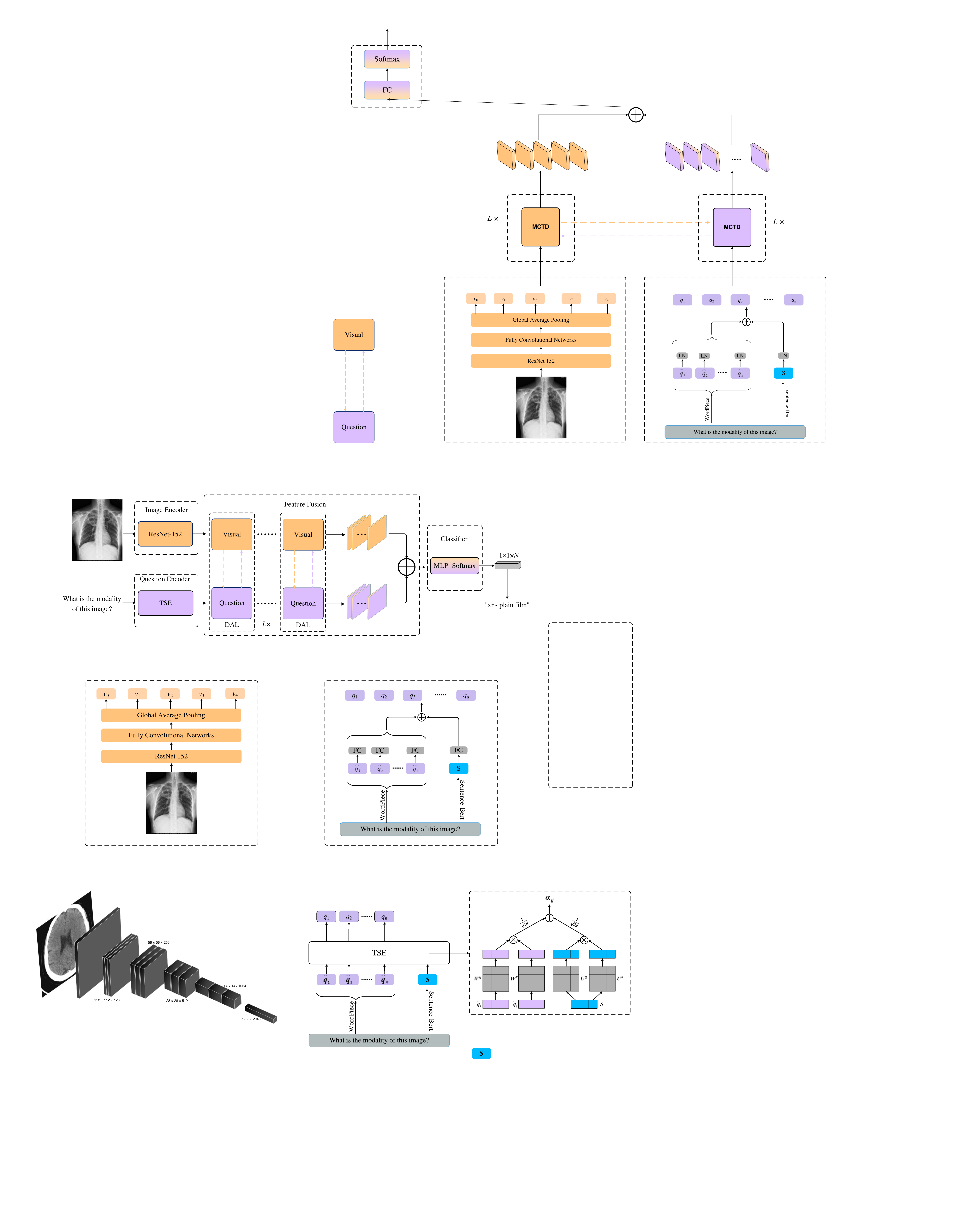}}
\caption{The proposed WSDAN framework. The ResNet-152 is used to extract image features, $V$. TSE is used to extract the question features, $Q$. The $L$ DALs form the fusion module, and the fusion vector is obtained with the help of a weighted combination. The classifier consists of \textit{MLP} and \textit{softmax}.}
\label{fig1}
\end{figure*}

\begin{figure}
\centerline{\includegraphics[width=\columnwidth]{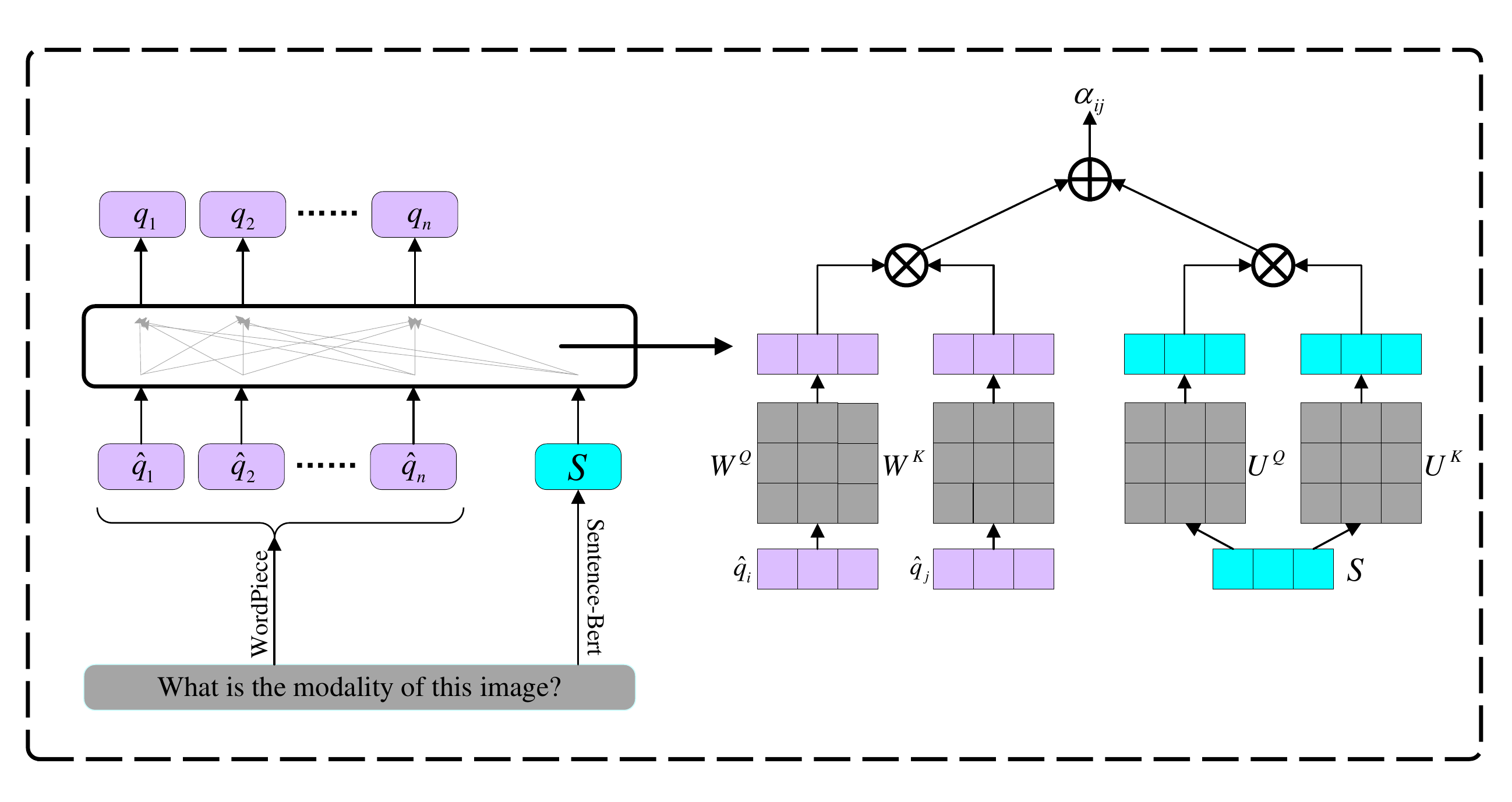}}
\caption{Question Encoder: TSE fuses word and sentence embedding to obtain a double embedding representation of the question.}
\label{fig2}
\end{figure}

\subsection{Fusion Algorithm}
The fusion phase involves modeling the correlation between the extracted visual and textual features. This phase mainly includes the attention mechanism and multi-modal pooling.

\subsubsection{Attention mechanism}
This mechanism has been widely used in the feature fusion phase of VQA and MVQA to help the models understand the visual content of images well. For example, Du \textit{et al.} used visual attention to understand visual features and find core image regions associated with question words for visual-text correlation learning\cite{deep}. However, in VQA, the model also needs to understand NL questions, so the model is required to have both visual and textual learning capabilities.

Lu \textit{et al.} constructed a co-attention learning framework called HieCoAtt, which alternates learning visual and textual attention in a hierarchical manner\cite{hiecoatt}. Nam \textit{et al.} proposed a combined framework for visual and textual attention learning, which draws on multiple steps to learn keywords and core regions\cite{dualat}. However, these schemes only learn separate attention distributions for each modality, ignoring the intensive interaction between each word and object. Liu \textit{et al.} proposed Cross-Attentional Spatio-Temporal Semantic Graph Networks (CASSG), a multi-headed, multi-hop attention model with diversity and progressivity, to explore fine-grained interactions among different modalities in an intersectional manner\cite{cassg}. Learning Cross-Modality Encoder Representations from Transformers (LXMERT)\cite{lxmert} and Vision-and-Language BERT (ViLBert)\cite{vilbert} learned about image and text co-attention with two different Transformer components, but no one has introduced these ideas into MVQA.

\subsubsection{Multi-modal pooling}
The fusion scheme of multimodal pooling is another common technique in VQA. The basic operations include concatenation, summation, and element-wise product, but these have mediocre performance.  Fukui \textit{et al.} proposed to embed image and text features into a higher-dimensional vector space to aggregate visual and linguistic features and developed Multimodal Compact Bilinear pooling (MCB), which greatly improved the integration of visual and textual elements\cite{mcb}. However, the memory usage problem caused by the high-dimensional output may limit its applicability. Multimodal Low-rank Bilinear Attention Networks (MLB) proposed by Kim \textit{et al.} decomposed the 3D weight tensor of a bilinear set into three 2D weight matrices, which solved the problem left by the MCB\cite{MLB}. But MLB is more complex and converges slowly. Yu \textit{et al.} proposed Multi-modal Factorized Bilinear (MFB) to sum the values within each non-overlapping 1D window and aggregate the results of multiplications between elements, thus increasing the model’s fine-grained fusion of vision and text\cite{mfb2}. 

\section{Method}
As with existing visual question answering methods, about the given medical image, $MV$, MVQA aims to predict the most reasonable and likely answer, $\hat{a}$, to a question, $MQ$. The task can be formulated as:
\begin{equation}\label{equ:PixelCCColorOrder}
\hat a=\mathop{\mathrm{argmax}}\limits_{a\in{\mathcal{A}}}P(a|MV,MQ,\theta)
\end{equation}
where ${\mathcal{A}}$ is the set containing all candidate answers, and $\theta$ represents all the parameters of the model.

The proposed WSDAN model framework is shown in \textcolor{mblue}{Fig. \ref{fig1}}. Specifically, we use the TSE module and pre-trained ResNet-152 to extract question features, $Q$, and image features, $V$, respectively. Through the fusion components, the understanding of the features is deepened and the visual reasoning ability is improved. The fusion vectors are then fed into a classifier that outputs the probability distribution of the $N=|{\mathcal{A}}|$ answers and obtains the predicted results. Next, we describe each step of the framework in detail.

\subsection{Image and Question Representations}
\subsubsection{Image representation}
Considering the limitations of the dataset and the goal to maximize the role of DAL in feature understanding, we choose the pre-trained ResNet-152 as the image encoder. As the medical images are extremely complex, similar to the CGMVQA approach, our proposed method extracts image features from five convolutional blocks of ResNet-152 to function as a detection model. On the one hand, it can make full use of image information, and on the other hand, it can help the model learn dual-attention better. The features of the given medical image $MV$ can be defined as $V=(v_1;v_2;\cdots;v_5)\in {\mathbb{R}}^{5\times d_v}$.

\subsubsection{Question representation}
The word information can ensure that the model selects the correct keywords for visual reasoning to predict the answer during feature understanding. However, using only word representations causes difficulty in understanding the given medical question correctly. Thus, we make up for this deficiency by learning the overall semantics of medical texts. We design the TSE module to obtain a double embedding representation of the question, ensuring that the extracted features are rich in both keywords and medical information. 

For the given question $MQ$, the word representation is formulated as follows. First, we use WordPiece to mark it as several words, then project it to the embedding layer to obtain the word representation, $\hat{q}_1,\hat {q}_2,...,\hat{q}_n$, where $\hat{q}_i\in{\mathbb{R}}^{d_q}$. The sentence representation is created by directly extracting the sentence embedding, $S=\mathrm{SBert}(MQ)\in{\mathbb{R}}^{d_q}$, for $MQ$ using pre-trained sentence BERT.

The attention mechanism can effectively fuse different levels of information. It is formulated as querying a dictionary with key-value pairs and can be reconstructed based on the similarity of the elements\cite{attention}, as in Equation (\ref{SA}):
\begin{equation}
\label{SA}
\mathrm{Attention}(Q,K,V)={\mathrm{softmax}}(\frac{QK^T}{\sqrt{d_k}})V
\end{equation}
where $d_k$ indicates the dimension of keys, $K$. Inspired by TUDE (Transformer with Untied Positional Encoding)\cite{TUDE}, we add sentence information to the relational modeling of each word pair to reduce the loss of semantic information. Specifically, we combine words and sentences embedding in the following way:
\begin{equation}
\alpha_{ij}=\frac{1}{\sqrt{2d}}(\hat{q}_iW^{Q})(\hat{q}_jW^K)^T+\frac{1}{\sqrt{2d}}(SU^{Q})(SU^K)^T
\end{equation}
\begin{equation}
q_i=\sum_{j=1}^{n}\frac{\exp(\alpha_{ij})}{\sum_{j'=1}^{n}\exp(\alpha_{ij'})}(\hat{q}_jW^V)
\end{equation}
where $W^Q, W^K, W^V\in\mathbb{R}^{d_q\times d_q}$ denote the learnable project matrices of $\hat{q}_i$. $U^Q, U^K\in\mathbb{R}^{d_q\times d_q}$ represent the learning project matrices of $S$, and $d_q$ is the dimension of the question feature space. Then, we obtain a double embedding representation of the question, $Q=(q_1;q_2;\cdots;q_n)=\mathrm{TSE}(MQ)\in\mathbb{R}^{n\times d_q}$. As shown in \textcolor{mblue}{Fig. \ref{fig2}}.

\subsection{Dual-Attention Learning}
Before describing feature fusion, we introduce its core component, DAL, which has an encoder-decoder architecture consisting mainly of self-attention and guided attention. DAL can learn simultaneously image- and question-guided attention. 

\subsubsection{Self-Attention and Guided Attention}
Multi-head attention based on the Equation (\ref{SA}) can provide information on the encoding of different subspaces and enhance the expressiveness of the model, which can be defined as:
\begin{equation}
\mathrm{MA}(Q,K,V) =[{\mathrm{head}}_1;{\mathrm{head}}_2;\cdots;{\mathrm{head}}_h]\mathbf{W}^o
\label{mulithead}
\end{equation}
\begin{equation}
{\mathrm{head}}_j =\mathrm{Attention}(Q\mathbf{W}^{Q}_j,K\mathbf{W}^{K}_j,V\mathbf{W}^{V}_j)
\end{equation}	
where $\mathbf{W}^{Q}_j,\mathbf{W}^{K}_j,\mathbf{W}^{V}_j\in\mathbb{R}^{d_q\times d_q}$ are learning matrices.

We build self-attention and guided attention based on the multi-head attention mechanism. Using the image and question encoder, we can extract the corresponding features from the input data and obtain image feature $V$ and question feature $Q$. The self-attention learning for element $v_i$ can be defined as $f_i=\mathrm{MA}(v_i,V,V)$, and reconstructed from the normalized similarity of $v_i$ and all samples in $V$. For the feature $V$, the self-attention learning can be expressed as:
\begin{equation}
F_{(V,V)}=\mathrm{MA}(V,V,V)
\label{sa}
\end{equation}

The guided attention learning for element $v_i$ can be expressed as $\textsl{g}_i=\mathrm{MA}(v_i,Q,Q)$, which is reconstructed by cross-modal similarity between $v_i$ and all samples in $Q$. The guided attention learning by $Q$ on $V$ is $F_{(V,Q)}$:
\begin{equation}
F_{(V,Q)}=MA(V,Q,Q)
\label{ga}
\end{equation}
We simulate intensive intramodal and intermodal interactions by learning self-attention and guided attention.

\subsubsection{DAL for MVQA}	

\begin{figure}[htbp]
\centering
\includegraphics[width=8cm]{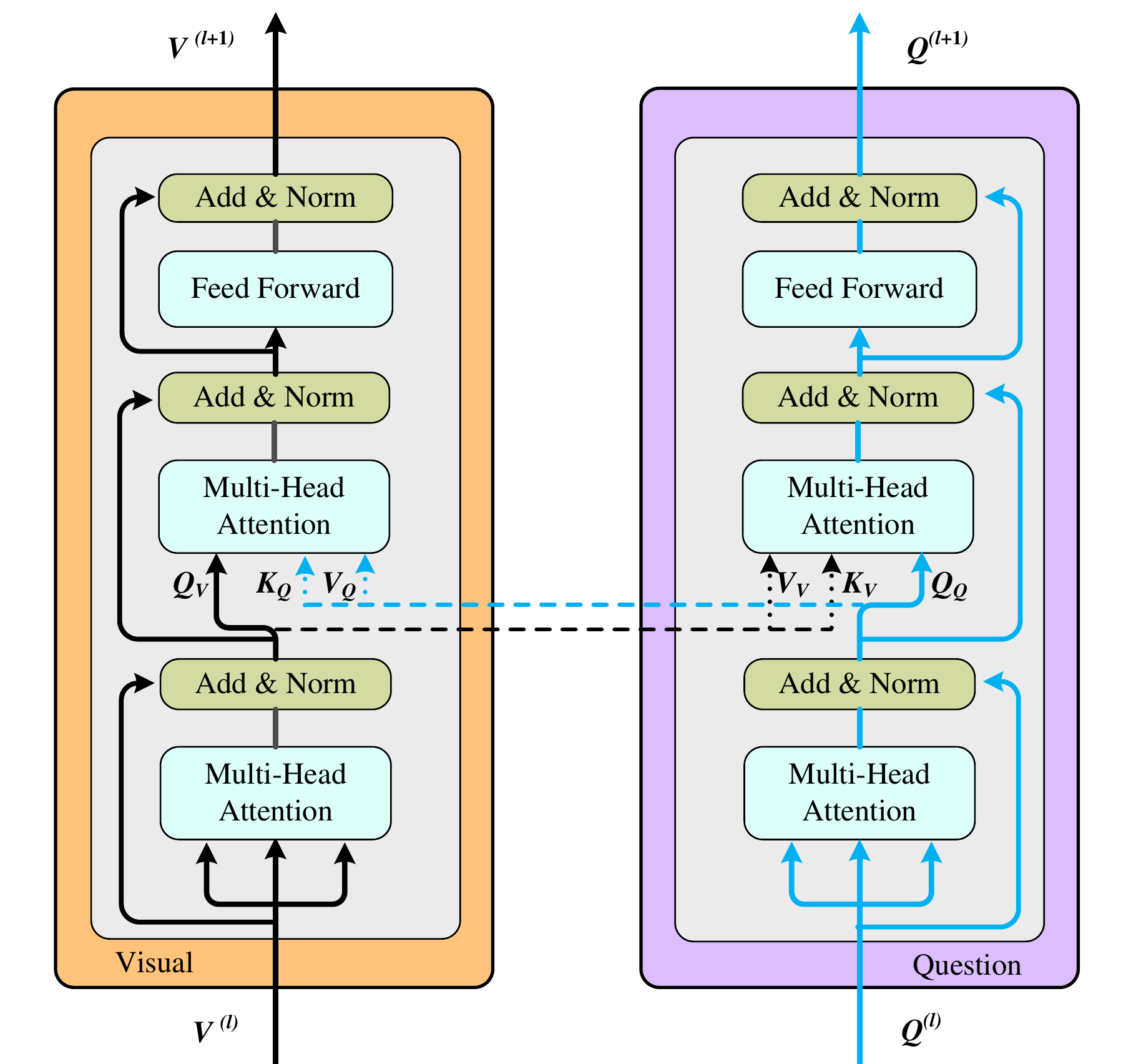}
\caption{Details of our proposed Dual-Attention Learning (DAL) module. Multi-head attention in the first and second layers is used to learn self-attention and guided attention, respectively.}
\label{fig4}
\end{figure}

The DAL module is shown in \textcolor{mblue}{Fig. \ref{fig4}}, we found that our DAL presents a bi-directional encoder-decoder structure. For example, on the left half of \textcolor{mblue}{Fig. \ref{fig4}}, this part can be considered as an encoder when learning image-guided attention and as a decoder when learning question-guided attention. Both the encoder and decoder in DAL have two layers of multi-head attention that learn self-attention and guided attention. We use the first part of multiple attention for self-attention learning and the second part for guided attention learning. Take visual learning as an example. First, the input image features, $V$, are subjected to self-attention learning of Equation (\ref{sa}), which models the interaction relationship between each pixel pair $\lbrace v_i,v_j\rbrace \in V$ and reconstructs $V$ based on the similarity between the pixels. Later, after guided attention learning of Equation (\ref{ga}), the interaction relationship between each $v_i\in V$ and $q_j\in Q$ is modeled, and $V$ is reconstructed again based on the similarity between the two modalities.

\subsection{DALs for Feature Understanding and Fusion}	
Modal information is initially fused during feature interaction and reconstruction. However, individual DAL plays a limited role, so we cascaded the DALs in depth. We learn co-attention between vision and text through the alternation of self-attention and guided attention in DALs, thus enhancing feature understanding while further facilitating the fusion of two forms of modal information. Specifically, the image feature, $V$, and question feature, $Q$, which are extracted by the image encoder and question encoder, are input to the DALs, denoted as $\mathrm{DAL}^{(1)},\mathrm{DAL}^{(2)},\cdots,\mathrm{DAL}^{(L)}$, where \textit{L} is the number of DAL; $V^{(l-1)}$ and $Q^{(l-1)}$ are the input of $\mathrm{DAL}^{(l)}$, the output features are represented by $V^{(l)}$, $Q^{(l)}$, which are further fed into the next DAL, namely, $\mathrm{DAL}^{(l+1)}$. The recursive process can be described as:
\begin{equation}
[V^{(l)},Q^{(l)}]=\mathrm{DAL}^{(l)}([V^{(l-1)},Q^{(l-1)}])
\end{equation}

In particular, we take $[V,Q]$ as the input feature of $\mathrm{DAL}^{(1)}$, i.e., $[V^{(0)},Q^{(0)}]=[V,Q]$. After DALs, $V^{(L)}=(v^{(L)}_1;\cdots;v^{(L)}_5)\in{\mathbb{R}}^ {5\times d_v}$ and $Q^{(L)}=(q^{(L)}_1;\cdots;q^{(L)}_n)\in{\mathbb{R} }^{n\times d_q}$ can be represented as the output image and question features. Then, we perform a simple weighted sum of $V^{(L)}$ and $Q^{(L)}$ as in Equation (\ref{weight sum}):
\begin{equation}
\label{weight sum}
\begin{aligned}
	Z = & \mathbf{W}^{V}V^{(L)}+\mathbf{W}^{Q}Q^{(L)}\\
	= & \sum_{i=1}^{5}\mathbf{w}^V_iv_i^{(L)} + \sum_{j=1}^{n}\mathbf{w}^Q_jq_j^{(L)}
\end{aligned}
\end{equation}
where $\mathbf{W}^V\in \mathbb{R}^{1\times 5}$, $\mathbf{W}^Q\in \mathbb{R}^{1\times n}$. $Z\in \mathbb{R}^{1\times d}$ is the ultimate fusion feature, which is then fed into a classifier consisting of MLP and softmax to obtain the most likely answer $h_c$.

\section{Experiments and Results}
In this section, we evaluate the performance of our proposed WSDAN on the datasets \textit{VQA-MED 2019} and \textit{VQA-RAD} in detail. In addition, we conduct ablation experiments to validate the effectiveness of the TSE and DALs, and use Gradient-weighted Class Activation Mapping (Grad-CAM) for visual analysis to explore the impact of using different guided attention on visual reasoning. We select accuracy and Bilingual Evaluation Understudy (BLEU) score as the evaluation metrics for the model.

\subsection{Datasets}
\textit{VQA-RAD}\cite{vqarad} is an MVQA dataset on radiology, which has 315 medical images and 3515 question–answer (QA) pairs of 11 types. Then, 58\% of the questions in \textit{VQA-RAD} are close-ended and the rest are open-ended. We train and test these two categories of data separately. \textit{VQA-RAD} is manually labeled and the trained model has a higher confidence level. However, this dataset only contains two parts, namely, training and testing, without a validation set.

\textit{VQA-MED 2019}\cite{vqamed} is presented in the ImageCLEF 2019 challenge. Inspired by \textit{VQA-RAD}, all questions in \textit{VQA-MED 2019} follow the patterns naturally proposed and validated in \textit{VQA-RAD}. The \textit{VQA-RAD} dataset covers the four most common categories in medical diagnosis: modality, plane, organ system, and abnormality. Of these, abnormality is open-ended, and the remaining categories are close-ended questions. We observed some “yes/no” types of questions in the modality and abnormality. To ensure comparability of experimental results and accommodate this sort of data, we created a new category called Yes/No. All QA pairs in this dataset are automatically generated by the algorithm from the image caption content, which contains a large amount of noise and useless information. Incidentally, both the \textit{VQA-RAD 2019} and \textit{VQA-MED} test results contain the overall category and are calculated according to the following rules:
\begin{equation}
{\rm{Overall}} = \sum_{i=1}^{M}C_i\frac{D_i}{T}
\end{equation}
where $C_i$, $D_i$ represents the result (accuracy or BLEU) and data volume of the \textit{i}-th category, respectively. \textit{M} denotes the number of categories, $M = 2$ in \textit{VQA-RAD} and $M=5$ in \textit{VQA-MED 2019}. \textit{T} represents the amount of data in the entire test set. We treat WSDAN as a classification model on both datasets.

Pretraining not only speeds up model convergence but also has the potential to achieve better results. To obtain the best performance WSDAN, we pretrain it on the \textit{Radiology Objects in Context} (\textit{ROCO}) dataset\cite{roco}. In the pretraining phase, we use all images and their corresponding captions and keywords for the Mask Language Model (MLM).

\subsection{Implementation details and training}
To improve the model and ensure that it has excellent generalization ability, we use cross-entropy with label smoothing as the loss function:
\begin{equation}
\begin{aligned}
	Loss=H(q',p)=&-\sum_{k=1}^{K}q'(k)\log p(k)\\
	=&(1-\epsilon)H(q,p)+\epsilon H(u,p)
\end{aligned}
\label{loss}
\end{equation}

where $p(k)$ means the probability that our model calculates each label $k\in \{1,...,K\}$, and $q'(k)=(1-\epsilon)q(k)+\epsilon u(k)$, $u(k)$ is a fixed distribution. During the experiment, we set $u(k)=1/K$ as a uniform distribution with respect to \textit{K}. $H(q,p)$ and $H(u,p)$ represent the cross-entropy loss when the true distributions are $q(k)$ and $u(k)$, respectively. $\epsilon$ represents the \textit{smoothing coefficient}.

In the pretrained MLM, the task is to predict tokens that are masked. Unlike unimodal interaction, in this task, we not only use the unmasked text but also incorporate visual information. To ensure that the model learns medical knowledge based on medical images and NL questions, we masked only the medical keywords provided in the original dataset.

The hyperparameters of WSDAN throughout the experiment are set as follows. The image feature dimension, $d_v$, question feature dimension, $d_q$, and hidden layer dimension, $d$, are equal to 312 (i.e., $d_v=d_q=d=312$), the number of heads in the multi-head attention layer $h=12$, the dimension of each head $d_h= d/h=26$, the uniform length of the padded text $n=20$, and the vocabulary size is 30,522. During the experiment, the number of DAL, $L=2$, and the \textit{smoothing parameter} $\epsilon=0.10$ in the loss function Equation (\ref{loss}). In the pretraining stage, $dropout=0.0$, and during finetuning, $dropout=0.1$.

We train our model on a single NVIDIA RTX 3080 GPU. We reshape all images to size $244\times 244$ and perform augmentation, including cropping, rotation, and color jitter, to enrich usable image information. In pretraining, we use the Adam optimizer to optimize the model loss with a learning rate of 2e-5, batch size=16, and iteration of 10 epochs. In finetuning, we adopt the Adam optimizer with a learning rate of 1e-4, batch size=16, and iteration of 100 epochs to optimize the model loss. In \textit{VQA-MED 2019} (or \textit{VQA-RAD}), if the loss on the validation set (or training set) does not improve within 10 consecutive epochs, the learning rate is reduced by a factor of 0.1. To avoid the influence of chance, we select the accuracy and BLEU score that appear frequently in the test results. The core code is available at \href{https://github.com/Coisini-Glenda/WSDAN-for-medical-visual-question-answering}{https://github.com/Coisini-Glenda/WSDAN}.

\subsection{Ablation Studies}

\begin{table*}[h]
\centering
\begin{center}
	\caption{Accuracy and BLEU score results of WSDAN with different states on \textit{VQA-MED 2019} dataset.}
	\label{tab4}
	\resizebox{\textwidth}{!}{
		\begin{tabular}{c|c|c|c|c|c|c|c|c|c|c|c|c}
			\toprule[1.25pt]
			\multirow{2}*{Method} & \multicolumn{2}{c|}{Modality} & \multicolumn{2}{c|}{Plane} & \multicolumn{2}{c|}{Organ} & \multicolumn{2}{c|}{Abnormality} & \multicolumn{2}{c|}{Yes/No} & \multicolumn{2}{c}{Overall}\\
			\cline{2-13}
			~&Acc&BLEU&Acc&BLEU&Acc&BLEU&Acc&BLEU&Acc&BLEU&Acc&BLEU\\
			\hline
			WDAN(NP) &81.9 & 86.2 &81.6 &81.6 &71.2 &74.6 &6.14 &7.92 & 81.2 &81.2 &63.2 &65.0\\
			\hline
			WSDAN(NP) &\textbf{83.3} &\textbf{87.9} &\textbf{83.2} &\textbf{83.2} &\textbf{72.8} &\textbf{75.9} &\textbf{12.3} &\textbf{13.4} &\textbf{82.8} &\textbf{82.8} &\textbf{64.5} &\textbf{66.2}\\
			\hline
			WSDAN(NP) only with $F_{(Q,V)}$  & 79.2 &84.1 &79.2 &79.2 &70.4 &72.9 &6.1 &6.3 &81.2 &81.2 &60.7 &62.1\\
			\hline
			WSDAN(NP) only with $F_{(V,Q)}$ &79.2 &84.2 &80.0 &80.0 &70.4 &75.2 &9.6 &11.3 &82.8 &82.8 &61.9 &64.2\\
			\bottomrule[1.25pt]
	\end{tabular}}
\end{center}
\end{table*}

We design ablation experiments to verify (a) whether the TSE module can effectively extract medical information from text, and (b) whether the proposed DALs can effectively enhance feature understanding and improve visual reasoning by simultaneously learning image- and question-guided attention. For these purposes, we set the following states: (1) WDAN(NP) has no pretraining and only applies word information as question features using both question- and image-guided attention models. (2) WSDAN(NP) has no pretraining and applies word and sentence information as question features using two types of guided attention. (3) WSDAN(NP) only with $F_{(Q,V)}$ represents a non-pretrained model that uses double embedding of words and sentences, but only uses image-guided attention. (4) WSDAN(NP) only with $F_{(V,Q)}$ has no pretrained model that uses double embedding but only uses question-guided attention. (5) WSDAN(P) is a pretrained model that uses double embedding and two types of guided attention.

\begin{table}[h]
\centering
\begin{center}
	\caption{Accuracy and BLEU score results of WSDAN with different states on \textit{VQA-RAD} dataset.}
	\label{tab5}
	\begin{tabular}{c|c|c|c}
		\toprule[1.25pt]
		\multirow{2}*{Method} & \multicolumn{3}{c}{VQA-RAD} \\
		\cline{2-4}
		~& Open-ended & Closed-ended & Overall \\
		\hline
		WDAN(NP) &57.5 & 76.5 &68.9\\
		\hline
		WSDAN(NP) &\textbf{63.7} &\textbf{80.1} &\textbf{73.5}\\
		\hline
		WSDAN(NP) only with $F_{(Q,V)}$  &59.8 &75.7 &69.4 \\
		\hline
		WSDAN(NP) only with $F_{(V,Q)}$ &60.9 &77.2 &70.7 \\
		\bottomrule[1.25pt]
	\end{tabular}
\end{center}
\end{table}

\subsubsection{Validation of TSE}\textcolor{mblue}{Table \ref{tab4}} and \textcolor{mblue}{Table \ref{tab5}} depict in detail the performance of WSDAN in different states on the \textit{VQA-MED 2019} and \textit{VQA-RAD} test sets, respectively. WSDAN(NP) is model with sentence semantics added to WDAN(NP). \textcolor{mblue}{Table \ref{tab4}} shows the following results. Compared with WDAN(NP), WSDAN(NP) outperformed WDAN(NP) significantly in all categories and exceeded it by approximately 10 percentage points in abnormality. As reported in \textcolor{mblue}{Table \ref{tab5}}, WSDAN(NP) has higher accuracy on the \textit{VQA-RAD} test set compared to WDAN(NP), improving to 63.7\% on the open-ended and 80.1\% on the close-ended questions. In summary, the performance of the model is significantly improved by adding sentence semantics to the use of word information. This result shows that the TSE module can be used to encode the input questions to obtain medical information-rich question features.

\subsubsection{Validation of DALs}We consider the impact of the use of different guided attention on WSDAN to verify whether the proposed DALs (learning both image- and question-guided attention) is the most effective. We analyze the differences in WSDAN performance using different guided attention according to \textcolor{mblue}{Table \ref{tab4}} and \textcolor{mblue}{Table \ref{tab5}}. Here, we compare the performance of WSDAN(NP), WSDAN(NP) only with $F_{(Q,V)}$, and WSDAN(NP) only with $F_{(V,Q)}$ on the two test sets. The results show that no matter which dataset is used,WSDAN(NP) only with $F_{(Q,V)}$ performs the worst, followed by WSDAN(NP) only with $F_{(V,Q)}$, and the best is WSDAN(NP). These comparative results show that DAL learning both types of guided attention can help the model to better understand visual and textual content. They also suggest that DALs, as shown in \textcolor{mblue}{Fig. \ref{fig4}}, is the most effective method.
\begin{figure}[h]
\centerline{\includegraphics[width=8cm]{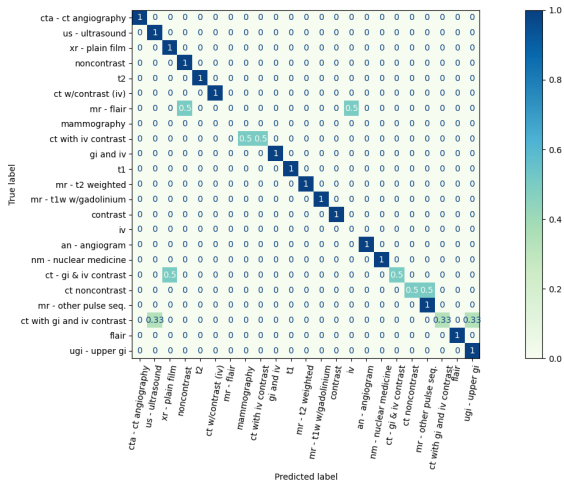}}
\caption{The confusion matrix of the Modality category.}
\label{fig7}
\end{figure}

\subsection{Qualitative Analysis}
We also use an additional evaluation matrix to assess the classification effect. We visualized the confusion matrices for the three categories in  \textit{VQA-MED 2019}: modality, plane, and organ. However, we did not visualize and analyze the confusing matrices for abnormality in \textit{VQA-MED 2019} and for the open-ended and close-ended categories in \textit{VQA-RAD} because there are too many candidate answers in these categories. The confusion matrix visualization of the modality category is shown in \textcolor{mblue}{Fig. \ref{fig7}}. Some types can be accurately predicted, such as "cta-ct angiography," "us-ultrasound," and others. However, none of the "Mr-flair," "mammography," and "iv" types were correctly predicted, and the model performed poorly among the remaining types.

\begin{figure}[h]
\centerline{\includegraphics[width=8cm]{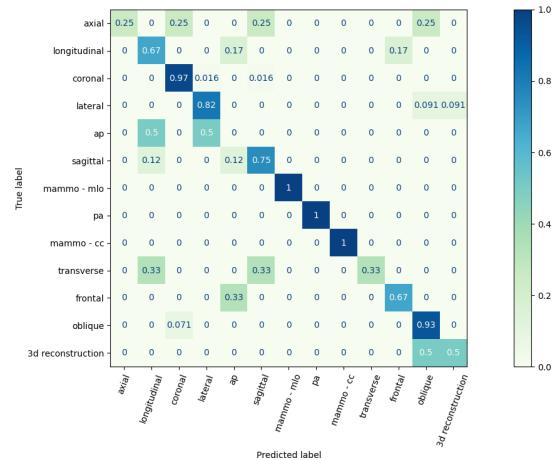}}
\caption{The confusion matrix of the Plane category.}
\label{fig8}
\end{figure}

A confusion matrix visualization of the plane category is shown in \textcolor{mblue}{Fig. \ref{fig8}}. In this category, "mammo-mlo," "pa," and "mammo-cc" can be completely predicted accurately, "ap" type is not predicted at all, "lateral," "conronal," "sagittal," and "oblique" can be predicted with high probability. However, it performs poorly in "axial," "transverse," and "3d reconstruction" and can only be correctly predicted with low probability.

\begin{figure}[h]
\centerline{\includegraphics[width=8cm]{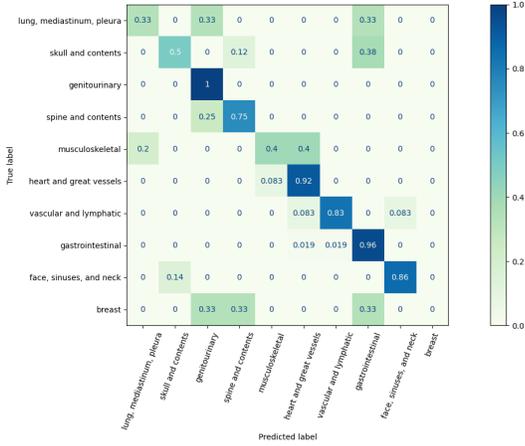}}
\caption{The confusion matrix of the Organ category}
\label{fig9}
\end{figure}	

\begin{figure}[h]
\centerline{\includegraphics[width=\columnwidth]{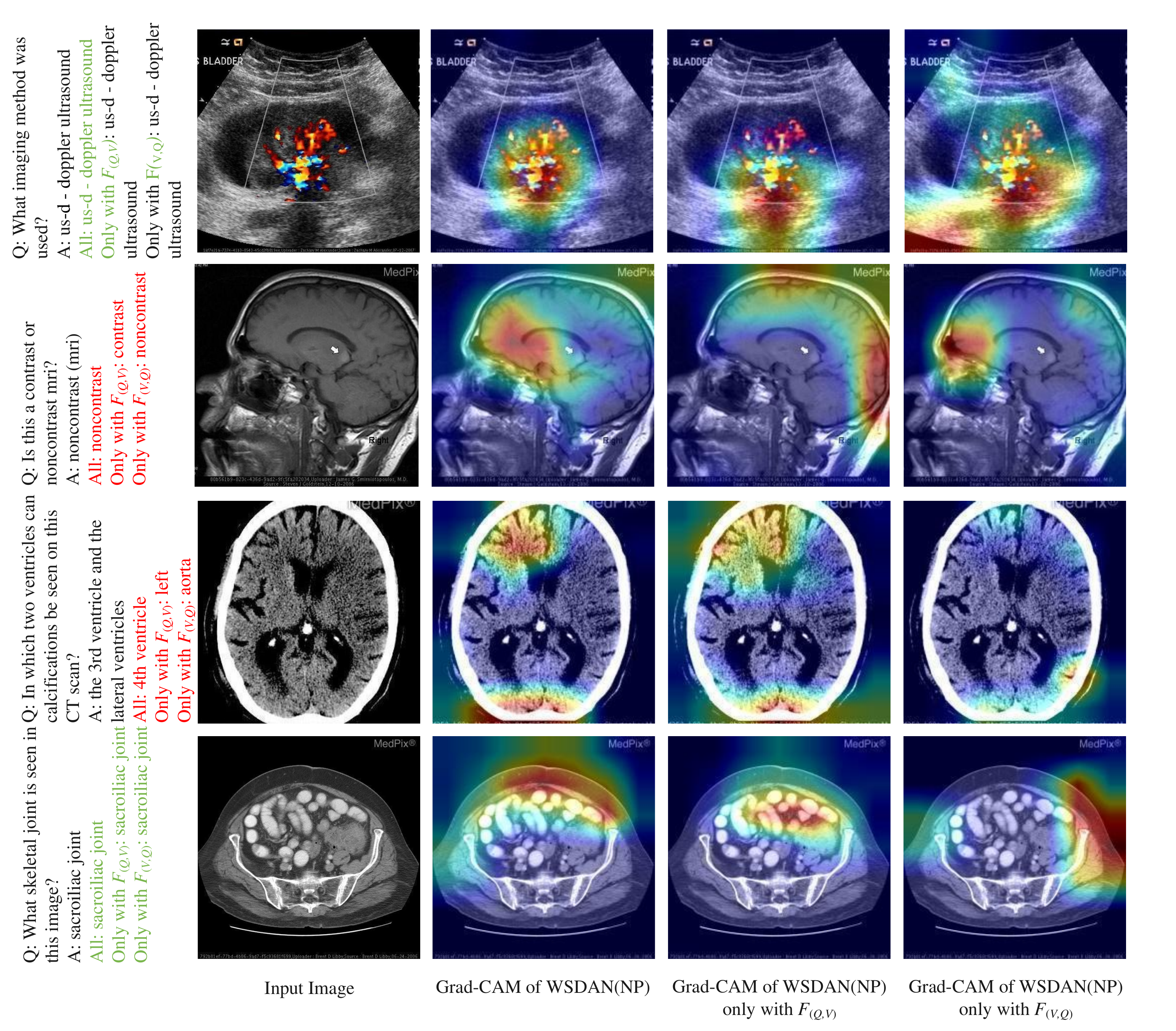}}
\caption{Example of medical images and Grad-CAM maps from WSDAN with different states on \textit{VQA-MED 2019} and \textit{VQA-RAD}. The text on the left shows the pairs of question-answer ground truths used in each row and the predictions of WSDAN(NP) (All), WSDAN(NP) only with $F_{(Q,V)}$ (Only with $F_{(Q,V)}$) and WSDAN(NP) only with $F_{(V,Q)}$ (Only with $F_{(V,Q)}$). Green and Red denote the correct and wrong predictions, respectively.}
\label{fig5}
\end{figure}

\textcolor{mblue}{Fig. \ref{fig9}} shows the confusion matrix visualization of the organ category. Among them, only the "genitourinary" type can be completely predicted, and most of them are correctly predicted with a high probability, such as "face, sinuses, and neck," "gastrointestinal," and "heart and great vessels," and others. "lung, mediastinum, pleura," "skull and contents," and "musculoskeletal" can only be predicted with small probability. Only "breast" cannot be predicted.

Then, with the aid of Grad-CAM, we analyze the effect of using different guided attention on visual reasoning. \textcolor{mblue}{Fig. \ref{fig5}} depicts four examples of Grad-CAM maps for different states of WSDAN in two datasets. On the far left are the selected QA pairs, the first column shows the input medical image, and the other columns show the activation feature maps visualized by Grad-CAM to illustrate the focus of WSDAN in the different states for the given examples. Specifically, the second column shows the Grad-CAM maps for WSDAN(NP), the third column shows the Grad-CAM maps for WSDAN(NP) only with $F_{(Q,V)}$, and the fourth column shows the Grad-cam maps for WSDAN(NP) only with $F_{(V,Q)}$.

The top two rows of \textcolor{mblue}{Fig. \ref{fig5}} illustrate examples from \textit{VQA-MED 2019}, both with closed-ended questions. As can be seen from its first row, predict the answers to the question "What imaging method was used?". WSDAN in all states can focus on the region of the image relevant to the question and correctly predict the answer. However, the WSDAN(NP) can focus on a more detailed area. 

As shown in the second line of \textcolor{mblue}{Fig. \ref{fig5}}, the question is "Is this contrast or noncontrast MRI?". Their Grad-CAM maps show that WSDAN(NP) and WSDAN(NP) only with $F_{(Q,V)}$ focus on the same region, both infer the answer "noncontrast". WSDAN(NP) only with $F_{(V,Q)}$ focuses on different areas, and "contrast" is the answer it predicts. Unfortunately, none of their predictions matched the correct answer, that is, "noncontrast (MRI)". In reality, however, we can tell the answer to this question by answering "noncontrast". This suggests that the visual reasoning of WSDAN(NP) and WSDAN(NP) only with $F_{(Q, V)}$ already meets real-world needs. This type of problem occurs mainly because the actual situation is not considered in the process of generating the labels. This is described again in the following discussion.

The last two rows of \textcolor{mblue}{Fig. \ref{fig5}} depicts Grad-CAM maps for two instances in \textit{VQA-RAD}, all open-ended questions. From the above analysis, we can observe from the model’s performance on the close-ended questions that WSDAN(NP) can focus well on the image regions relevant to the questions. However, when faced with open-ended questions, the model is slightly inadequate. For example, in response to the question "In which two ventricles can calcifications be seen on this CT scan?," WSDAN failed to focus on the correct location in all states, predicting wrong answers and even some answers that were unrelated to the question (e.g., WSDAN(NP) only with $F_{(V,Q)}$ answered "aorta"). In response to the question "What skeletal joint is seen in this image?," although the correct response was predicted, the region of interest of the model was not relevant to the question.

In summary, our WSDAN(NP) can enhance the model's understanding of features and improve visual reasoning on closed questions by alternating learning between image- and question-guided attention and self-attention. On open-ended questions, the visual inference demonstrated is weak although the test performed better than the relevant models.

\begin{table*}[h]
\centering  
\begin{center}
	\caption{Accuracy and BLEU score test results of WSDAN(P) and related models on the \textit{VQA-MED 2019} Dataset. These are the results that come up frequently during  testing and can be used to represent the performance of WSDAN. NP and Ens refer to non-pretrained and ensemble models respectively. P means pretrained model.}
	\label{tab2}
	\resizebox{\textwidth}{!}{
		\begin{tabular}{c|c|c|c|c|c|c|c|c|c|c|c|c}
			\toprule[1.25pt]
			\multirow{2}*{Method} & \multicolumn{2}{c|}{Modality} & \multicolumn{2}{c|}{Plane} & \multicolumn{2}{c|}{Organ} & \multicolumn{2}{c|}{Abnormality} & \multicolumn{2}{c|}{Yes/No} & \multicolumn{2}{c}{Overall}\\
			\cline{2-13}
			~&Acc&BLEU&Acc&BLEU&Acc&BLEU&Acc&BLEU&Acc&BLEU&Acc&BLEU\\
			\hline
			TUA1\cite{tua1} &66.7 &66.7 &71.6 &83.4 &74.4 &75.1 &3.5 &8.8 &78.1 &78.1 &60.6 &63.4\\
			Up-Down &80.6 &87.1 &82.4 &83.4 &74.4 &75.1 &- &- &71.9 &71.9 &- &-\\
			CGMVQA &80.5 &85.6 &80.8 &81.3 &72.8 &76.9 &1.7 &1.7 &75.0 &75.0 &60.0 &61.9 \\
			CGMVQA.Ens &81.9 &88.0 &\textbf{86.4} &\textbf{86.4} &\textbf{78.4} &79.7 &4.40 &7.60 &78.1 &78.1 &64.0 &65.9\\
			MMBERT(NP) &80.6 &85.6 &81.6 &81.6 &71.2 &74.4 &4.30 &5.70 &78.1 &78.1 &60.2 &62.7\\
			MMBERT(P) &83.3 &86.2 &\bf{86.4} &\bf{86.4} &76.8 &80.7 &14.0 &16.0 &87.5 &87.5 &67.2 &69.0\\
			\hline
			\textbf{WSDAN(P)} &\textbf{84.7} &\textbf{89.8} &\textbf{86.4} &\textbf{86.4} &76.8 &\textbf{80.9} &\textbf{20.1} &\textbf{21.8} &\textbf{89.0} &\textbf{89.0} &\textbf{69.0} &\textbf{71.1}\\
			\bottomrule[1.25pt]
	\end{tabular}}
\end{center}
\end{table*}

\subsection{Comparison with State-of-the-Art}	
After conducting ablation studies, we compare the best single model WSDAN(P) with several state-of-the-art methods currently on the \textit{VQA-MED 2019} and \textit{VQA-RAD} datasets. The performance of WSDAN(P) on \textit{VQA-MED 2019} and \textit{VQA-RAD} is shown in \textcolor{mblue}{Table \ref{tab2}}, \textcolor{mblue}{Table \ref{tab3}}.

\begin{table}[h]
\centering
\begin{center}
	\caption{Accuracy of WSDAN(P) and related models on the \textit{VQA-RAD} test set. These are frequent results during testing and can be used to represent the performance of WSDAN.}
	\label{tab3}
	\begin{tabular}{c|c|c|c}
		\toprule[1.25pt]
		\multirow{2}*{Method} & \multicolumn{3}{c}{VQA-RAD} \\
		\cline{2-4}
		~& Open-ended & Closed-ended & Overall\\
		\hline
		BiAN & 28.4 & 67.9 & 52.3\\
		MAML(BAN) & 40.1 & 72.4 & 59.6\\
		MEVF(BAN) & 43.9 & 75.1 & 62.7\\
		MMQ\cite{MMQ} & 53.7 & 75.8 & 67.0\\
		CR\cite{CR} & 60.0 & 79.3 & 71.6 \\
		MMBERT(P) & 63.1 & 77.9 & 72.0\\
		CPRD+BAN+CR\cite{Cprd}  &61.1 &80.4 &72.7\\
		BiLR&\textbf{66.48} &82.35&76.05\\
		\hline 
		\textbf{WSDAN(P)} &65.9 & \textbf{83.8} & \textbf{76.69}\\
		\bottomrule[1.25pt]
	\end{tabular}
\end{center}
\end{table}

We describe the performance of WSDAN(P) on \textit{VQA-MED 2019} in detail. We find that the performance on several categories is significantly better than that of the previous correlation model, especially in abnormality. And the accuracy and BLEU score in the overall category improved by 1.8 and 2.1 percentage points to 69.0\% and 71.1\%, respectively. 

However, WSDAN(P) performed the worst in the organ category of \textit{VQA-MED 2019}. We attempted to guess the possible reasons, which is mainly an issue with the true label distribution of the dataset. We carefully analyzed the relevant data for the organ category and found that in the training and validation sets, "lung, mediastinum, pleura," "skull and contents," "genitourinary," "musculoskeletal," and others are the 10 types of answers. However, in the test set, another seven types include "gastrointestinal\#lung, mediastinum, pleura," "heart and great vessels\#lung, mediastinum, pleura\#spine and contents," and others, which never appeared during training and accounted for 7.2\% of the test set. In other words, the models had never been trained with these types of data, so were not aware of the visual and textual features corresponding to these answers. Thus, all the models performed poorly in the organ category. The result suggests that this dataset needs further refinement.

In \textit{VQA-RAD}, the accuracy of WSDAN(P) improved to 83.8\% on the closed-ended questions and 76.69\% in the overall category, outperforming all relevant models. The accuracy on the open-ended question is 65.9\%, which is about 1 percentage point lower than BiLR's 66.48\%. Perhaps the advantages of WSDAN on \textit{VQA-RAD} are not obvious, but while most of the existing work only considers the performance of the respective models on a single dataset, we consider two datasets and achieve better performance on both of them. This result suggests that our proposed WSDAN is more general and effective, allowing for more robust visual inference.

\section{Discussion and Conclusion}

The images and texts given in the MVQA task are more difficult to understand than normal because of the knowledge associated with the medical field. Extracting medical information from the text and making a fine-grained understanding of the image and question features is key to improving model prediction performance. The medical images in the \textit{VQA-MED 2019} and \textit{VQA-RAD} datasets cover almost all organs of the human body, and the corresponding natural language questions are the most common when medical images are read by physicians. In contrast to methods proposed in previous studies, WSDAN is effective on both datasets.

With regard to extracting richer information from medical natural language questions, most previous works only considered information about words or word combinations. In contrast to these schemes, WSDAN considers sentence semantics, aggregate words, and sentence information through TSE to extract text features that are rich in keywords and medical information. This idea may be applied to other related fields, such as biomedical text classification, which we believe could be highly interesting.

In this study, we tried to combine self-attention and different guided attention to fuse modal information, and the results show that using both image- and question-guided attention models performs best. Although more effective than previous fusion mechanisms, the classification model-based image encoder limits the play of DAL. We expect to interact image objects with question keywords to fully exploit the use of guided attention. In other words, we urgently need an image detection model to act as an image encoder for MVQA.

In this paper, we proposed WSDAN, a new model for MVQA tasks to efficiently extract medical information from natural language questions and explore the role relationship between vision and text. We designed a TSE module to add sentence information to each pair of word relationship models with the help of learnable projection matrices. This module ensures that the extracted question features are rich in keywords and medical information. The proposed DAL module effectively models intramodal and intermodal interactions through self- and guided attention learning. DALs can improve the model’s fine-grained understanding of features and enhance visual reasoning by learning co-attention between vision and text. We conducted comprehensive experiments on \textit{VQA-MED 2019} and \textit{VQA-RAD} datasets to confirm the effectiveness and generality of WSDAN. We believe that with further research, our approach can perform better and facilitate the development of CAD.

{\small
\bibliographystyle{IEEEtran}
\bibliography{IEEEabrv,IEEEtrans}
}

\end{document}